# LADDER: A Human-Level Bidding Agent for Large-Scale Real-Time Online Auctions

**Yu Wang, Jiayi Liu, Yuxiang Liu, Jun Hao, Yang He, Jinghe Hu, Weipeng P. Yan, Mantian Li**

Business Growth Division. JD.com.
Beijing, China
{wangyu5, liujiayi5, liuyuxiang1, haojun, landy, hujinghe, paul.yan, limantian}@jd.com

**Abstract**

We present LADDER, the first deep reinforcement learning agent that can successfully learn control policies for large-scale real-world problems directly from raw inputs composed of high-level semantic information. The agent is based on an asynchronous stochastic variant of DQN (Deep Q Network) named DASQN. The inputs of the agent are plain-text descriptions of states of a game of incomplete information, i.e. real-time large scale online auctions, and the rewards are auction profits of very large scale. We apply the agent to an essential portion of JD's online RTB (real-time bidding) advertising business and find that it easily beats the former state-of-the-art bidding policy that had been carefully engineered and calibrated by human experts: during JD.com's June 18th anniversary sale, the agent increased the company's ads revenue from the portion by more than 50%, while the advertisers' ROI (return on investment) also improved significantly.

## Introduction

Researchers have made great progress recently in learning to control agents directly from raw high-dimensional sensory inputs like vision in domains such as Atari 2600 games (Mnih et al. 2015), where reinforcement learning (RL) agents have human-level performance. However, most real-world problems have high-level semantic information inputs rather than sensory inputs, where what human experts usually do is to read and understand inputs in plain-text form and act after judging by expertise. Real-world problems are much more challenging than video games in that they always have a larger solution space and in that their states can only be partially observed. Such real-world problems have not been tackled by any state-of-the-art RL agents until now.

This paper demonstrates an agent named LADDER for such a problem. Using a deep asynchronous stochastic Q-network (DASQN), the agent improves the performance of JD's real-time bidding (RTB) ad business.

RTB is the most promising field in online advertising which greatly promotes the effectiveness of the industry (Yuan, Wang, and Zhao 2013). A typical RTB environment (Figure 1) consists of ad exchanges (ADXs), supply side platforms (SSPs), data management platforms (DMPs) and demand side platforms (DSPs). ADXs and DSPs utilize algorithms to buy/sell ads in real-time. SSPs integrate information of publishers (i.e. online media) and offer ads requests of the publishers to ADXs. An ADX puts the offers out to DSPs for bidding. DSPs target appropriate ads to the involved user based on information supplied by DMPs and return the ads with their bids to the ADX which displays ads of the highest bidder and charges the winner DSP with general second price (Varian 2007).

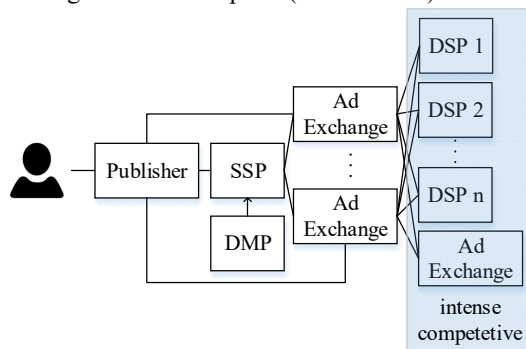

*Figure 1: A typical RTB auction environment*

Obviously, the process of many DSPs/ADXs bidding for an ad offer is an auction game (Myerson 1981) of incomplete information. However, the online ads industry just ignores this fact and considers RTB a solved problem: all existing DSPs model auction games as supervised learning (SL) problems by predicting the click through rate (CTR) (McMahan et al. 2013) or conversion rate (CVR) (Yuan, Wang, and Zhao 2013) of ads and using effective cost per mille (ECPM) as bids (Chen et al. 2011).

JD.com started its DSP business in 2014, at first we employed the industry state-of-the-art approach of ECPM bidding with a calibrated CTR model (McMahan et al. 2013) as depicted in Figure 2. Soon we found it impossible for the SL calibration model to have a stable performance in practice, which was critical for the business to keep breaking even. As a result, we introduced a method with fine grained bid coefficients calibrated by human experts. In a nutshell, our bidding mechanism then was a human-machine hybrid control system where operators modified the calibration coefficients tens of times per day.

For the obvious inefficiency of the hybrid system, we started research on utilizing RL algorithms to solve the auction game, during which we met several problems:

First, the solution space of the auction game is tremendous. JD DSP system is bidding for 100,000s of auctions per second, assume we have 10 actions and each day is an episode (ad plans are usually on a daily basis), simple math shows the solution space is of $10^{10^9}$. For comparison, the solution space of the game of Go is about $10^{360}$ (Allis and others 1994; Silver et al. 2016).

Second, state-of-the-art RL algorithms are inherently sequential, hence cannot be applied to large-scale practical problems such as the auction game, for our online service cannot afford the inefficiencies of sequential algorithms.

Third, auction requests are actually triggered by JD users and randomness of human behaviors implies stochastic transitions of states. That's very different from Atari games, text-based games (Narasimhan, Kulkarni, and Barzilay 2015) and the game of Go (Silver et al. 2016).

Besides, we have widely ranged rewards of which the maximum may be 100,000 times larger than the minimum, which implies only very expressive models are suitable.

Last but not least, there's much human-readable high level semantic information in JD which is crucial for bidding, e.g. the stock keeping units (SKUs) that a customer viewed or bought recently, how long ago she viewed or bought them, the price of the advertised SKU, etc. Although sophisticated feature engineering can utilize these information in a model like wide and deep models (Cheng et al. 2016) or factorization machines (Rendle 2012) as is already in place in the hybrid system, taking into account JD's scale, such models will be of billions of features and therefore too heavy to react instantly to the rapidly varying auction environment, leading to poor performance.

In this paper, we model the auction game as a partially observable Markov decision process (POMDP) and present the DASQN algorithm which successfully solve the inherently synchronousness of RL algorithms and the stochastic transitions of the game. We encode each auction request into plain text in a domain specific natural language, feed the encoded request to a deep convolutional neural networks (CNN) and make full use of the high-level semantic information without any sophisticated feature engineering. This results in a lightweight model both responsive and expressive which can update in real-time and reacts to the changes of the auction environment rapidly. Our whole architecture is named LADDER.

We evaluated LADDER on a significant portion of JD DSP business with online A/B test and the experimental results indicate that the industry was far from solving the RTB problem: LADDER easily outperformed the human expert calibrated ECPM policy: during JD.com's June 18th anniversary sale, the agent raised the company's ads revenue from the portion by more than 50%, while the ROI of the advertisers also improved as much as 17%.

## Background

RL provides the ability for an agent to learn from interactions with an environment $E$. In this paper, we consider the auction environment as $E$. At each time step $t$, the agent observes an auction $x_t$ from a publisher and a set of ads will participate in the auction. The agent selects a legal action $a_t \in A = \{0, ..., C\}$ and acts in $E$. After a while, the agent gets a real number reward $r_t$ from $E$. We formularize this sequential process as $(x_0, a_0, r_0, ..., x_t, a_t, r_t, ...)$ whose dynamics can be defined by the joint probability distribution $\Pr\{X_{t+1} = x_{t+1}, R_t = r_t \mid x_0, a_0, r_0, ..., x_t, a_t\}$.

Obviously $x$ cannot fully reflect the state of $E$. In fact, we define the state of $E$ at $t$ as $s_t = \phi(x_0, a_0, r_0, ..., x_t)$. $s_{t+1}$ depends on $s_t$ and $a_t$ with a certain probability. We define dynamics of $E$ as $p(s_{t+1}, r_t | s_t, a_t) = \Pr\{S_{t+1} = s_{t+1}, R_t = r_t | S_t = s_t, A_t = a_t\}$.

We model the auction game as a POMDP rather than a standard MDP because in such a real-world problem very little of the state can be observed (e.g. we never know the users' behaviors in physical stores). The game is assumed to terminate and restart in cycle. The state space of the POMDP is huge but still finite, standard RL methods such as Q-learning or policy gradient can be applied to learn an agent through the interaction with $E$.

Q-learning and its variants especially DQN (Mnih et al. 2015) learns a value function $Q(s, a; \theta)$ which indicates the future rewards since current state and derives a policy $\pi^*(s, a) = argmax_a Q(s, a; \theta), a \in \{1, 2, ..., C\}$. The loss function of DQN at step $t$ is defined as:

$$L(\theta_t) = (r_t + \gamma max_a Q(s_{t+1}, a; \theta_T) - Q(s_t, a_t; \theta_t))^2 \quad (1)$$

where $s_{t+1}$ is the state next to $s_t$ and $\theta_T$ keeps a periodic copy of $\theta$. $L(\theta_t)$ is the foundation of our formulation.

## Related Work

(Mnih et al. 2015) proposed DQN which combined RL and CNN and learned directly from screen pixels and outperformed human experts in Atari 2600 games. (Gu et al. 2016) improved the method with a new network architec-ture. (Van Hasselt, Guez, and Silver 2016) proposed Double DQN to tackle the overestimate problem in DQN.

POMDPs were well studied in (Jaakkola, Singh, and Jordan 1995), (Kaelbling, Littman, and Cassandra 1998), and (Monahan 1982). (Hausknecht and Stone 2015) modeled Atari 2600 games as POMDPs by replacing a full connection lay in DQN by an LSTM.

All these algorithms are sequential in that they can only act once after each step of SGD, which is unacceptable in our application scenario. (Mnih et al. 2016) presented A3C and n-step Q-learning among other asynchronous algorithms which decoupled RL algorithms to some extent in that agents could act n steps between 2 training steps, as well as could learn from several copies of the same game at the same time. However, A3C and n-step Q-learning still cannot solve the auction game because our scale requires full decoupling rather than semi decoupling.

In parallel with our work, (Cai et al. 2017) presented a RL method for RTB problem based on dynamic programming and CTR prediction, which also went beyond the traditional ECPM policy.

(Silver et al. 2016) applied RL, CNNs and Monte Carlo Tree Search to the game of Go and their agent namely AlphaGo beat the top human experts in an open competition. We argue that our auction game has a much larger solution space than Go, which makes tree search methods thoroughly impractical. Furthermore, Go is of perfect information, while auction games are of incomplete information in the form of human readable high-level semantic info.

Recurrent neural networks, especially LSTMs are extensively used in NLP tasks. Text-based game was researched by (Narasimhan, Kulkarni, and Barzilay 2015), they used LSTMs instead of the CNN in DQN. However, the two game studied had a tiny state space compared to auction games. In addition, RNNs need very sophisticated feature engineering to understand high-level semantic information, which makes the model too large to react instantly.

Character-level CNNs were proposed by (Zhang, Zhao, and LeCun 2015), which perform well on text classification tasks without word embedding. (Kim et al. 2016) introduced another character-level CNN with character embedding as inputs.

## The Architecture of the DSP System in JD

As the largest retailer in China, JD.com started its DSP business as early as 2014 to satisfy merchants' increasing demands for more sales. An overview of the architecture of our DSP system is illustrated in Figure 2. When an auction request from an ADX arrives, the system recalls hundreds of ads inventories as candidates from an ads repository with millions of ads. The ranking module ranks these candidates and identifies the top few ads for bidding (typically top 1). The bidding module computes and returns the ads and bid to the ADX as described in the induction section.

The industrially proven auction mechanism in such auction games is the general second price (GSP) method which has a Nash equilibrium in position auctions and is extensively used all over the world. In a GSP auction, a winner DSP knows only the bid of the DSP in the place immediately behind it because that's the winner's charge, but none of the losers knows anything about any rivals' bid. DSPs don't even know how many rivals are bidding in the auction. The auction game is a typical game of incomplete information where each DSP is a player (Gibbons 1992).

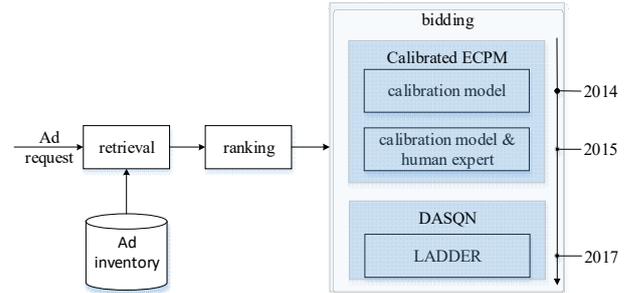

*Figure 2: The design and evolution of JD DSP's architecture*

The universal business mode of DSPs is that ad impressions from ADX are bought by cost per mille (CPM) and sold to advertisers by cost per click/action (CPC/CPA) to maximize ads performance. Though JD has several charging mechanisms other than CPC (CPA, for example), we speak of CPC in this paper for simplicity and the methods discussed are applicable to others.

We used ECPM = $Q * \text{bid}_{click}$ as described in (Varian 2007) for ranking, in which Q reflects business requirements (e.g. predicted CTR/CVR of the ads) and $\text{bid}_{click}$ was advertisers' CPC bids for their clicks. So there is a natural gap between revenue and expenditure which we must control in the bidding module.

Since 2014, we were using the state-of-the-art ECPM bidding policy. We tried to calibrate Q to a click through rate ($\text{CTR}_{FM}$) as depicted in (McMahan et al. 2013), except that we used a factorization machine (Rendle 2012) instead of Poisson regression for calibration. Our ranking model has a structure similar to wide and deep models (Cheng et al. 2016) with billions of weights and tens of gigabyte of memory and disk space requirements, meaning Q can hardly react to the rapidly changing auction environment without delay because the model is too huge to update in time, leading us to design a real-time impression-click data stream for online learning of the calibration CTR model. Afterwards the data stream was reused by LADDER.

Handling the huge amount of SKUs and hundreds of millions of active users of JD and tens of unknown rival DSPs exceeded the system's capabilities. Moreover, business requirements demand tradeoffs between profits and total revenue, e.g. maximize revenue while keeping certain net profit margins to generate economies of scale. To fulfill such requirements, at the end of 2015 we introduced a mechanism with traffic-type level coefficients of bid calibrated by human experts.

Consequently, the human-machine hybrid control system computed the bid of every auction as $\text{bid}_{DSP} = \text{Coef}_{human} * \text{CTR}_{FM} * \text{bid}_{cost/click}$ where human experts modified $\text{Coef}_{human}$ tens of times per day.

### The Learning Ad Exchange Bidder

In early 2016, we began the research of applying RL algorithms to the RTB auction games. Finally we succeeded in devising a RL agent named LADDER (short for the **l**earning **ad** exchange bi**dder**).

### Modeling

We model the auction game as a POMDP. Here we give some important definitions about the POMDP.

*Episodes*. Naturally, we define every day as an episode.

*Rewards*. To control deficits, we use net profits of every auction as rewards of LADDER. Assume our expense (by CPM) and income (by CPC) at time $t$ is $i_t$ and $e_t$ respectively, the reward of the auction at time $t$ is $r_t = i_t - e_t$. For simplicity we use CNY as units of all related variables. Notice that $i_t$ is always zero unless the user click the ad.

In practice, non-zero $i_t$ is usually $10^3 \sim 10^5$ times larger than $e_t$. Considering the relatively low click rate, the function we are fitting is extremely steep with most of its values negative while a small subset are high positive. To avoid financial loss, both the tiny negative values and the positive ones must be caught exactly by our model. This implies that high expressive models as CNNs are required.

*Actions*. We define actions of the auction game at time $t$ as $a_t = bid_t$ because bids happen to be discrete. Assume our bid ceiling is $C$, our actions would be from the set $A = [0, 0.01, 0.02, ..., C]$ because the minimal unit of CNY is $0.01$. As a result, our action space is in thousands and expected to be very sparse in the training data.

*States*. The high-level semantic information $Info_t$ we can get in JD at time $t$ is about active users, SKUs and ads. That's the partially observable state. Generally, the $t$th auction can be formularized as a text description in a domain specific natural language according to $Info_t$, as shown in the following example. All high-level semantic information in the example is in *italic*:

> Here's an auction from publisher *p*: user *u* is accessing *some_site.com/p*, *u* has bought SKUs of ID *s1*, *s2* and *s3 a* days ago, *u* browsed SKUs of ID *s4* and *s5 b* days ago… The candidate ad is SKU *s6* which is delivered by JD logistic network...

Notice that all the numbers above (*s1… s6, a, b*) are in plain text. There is a practical reason: ID numbering rule of JD requires that similar entities have close IDs, e.g. iphone7's ID is *3133817*, and iphone7 plus's ID is *3133857* which seems similar, so an experienced expert can judge from plain text that *3133857* would have similar performance as *3133817* in the same auction context even if she's never seen the former. RNN-based NLP models need elaborate feature engineering (e.g. character *n*-grams) to utilize such semantic, but such models will comprise billions of weights and therefore too large to react instantly to the auction environment, as discussed earlier. On the contrary, CNNs are good at recognizing similar patterns.

Based on this interesting observation as well as the definitions, we manage to build a solution. For the $t$th auction, we have a function $\phi$ which generates a text description from $Info_t$ as the above example and one-hot encodes the text as described in (Zhang, Zhao, and LeCun 2015) and feeds the encoded content to a CNN. In fact, the model works well without elaborate feature engineering, thus is space-efficient enough (less than 1Mb) to update instantly.

| Layer | Input | Patch size | Weights |
|---|---|---|---|
| Convolution | $600 \times 71 \times 1$ | $71 \times 7$ | 9.94K |
| Max pool | $594 \times 1 \times 20$ | $3 \times 1$ | |
| Convolution | $198 \times 1 \times 20$ | $7 \times 1$ | 2.8K |
| Max pool | $192 \times 1 \times 20$ | $3 \times 1$ | |
| Convolution | $64 \times 1 \times 20$ | $5 \times 1$ | 5.0K |
| Max pool | $60 \times 1 \times 50$ | $3 \times 1$ | |
| Convolution | $20 \times 1 \times 50$ | $5 \times 1$ | 12.5K |
| Max pool | $16 \times 1 \times 50$ | $2 \times 1$ | |
| Hidden | $8 \times 1 \times 50$ | | 160K |
| Linear | 400 | | |

*Table 1: Architecture of LADDER's convolutional network*

In our productive model, the input text is encoded into a $600 \times 71$ matrix, of which $600$ is the max length of the description and $71$ is the alphabet size. In order to save response time of the online service, we formulize the input text in a sort of shorthand with only key information rather than in full text. Also, we use a traditional architecture rather than the state-of-the-art Inception networks or ResNets (Szegedy et al. 2017) for the same reason.

Table 1 depicts the architecture of our model with output number of the linear layer (aka action space and bid ceiling) omitted deliberately for commercial privacy. All layers except the last one use RELU as activation functions.

### Deep Asynchronous Stochastic Q-learning

RL algorithms are inherently sequential, though A3C and other algorithms in (Mnih et al. 2016) made it possible to act an entire episode between each training step, they are still sequential in nature because the two processes of acting and training in those algorithms are still serially exe-cuted. That's unacceptable for an online DSP service that must respond to each of the huge amount of auctions in several milliseconds. From this perspective, training during serving is absolutely unfeasible, needless to say it requires hundreds of times more servers, which is uneconomical.

Distinguishingly, we solve this problem by introducing a fully decoupled parallel mechanism which results in a fully asynchronous RL algorithm in which all three processes (learning from the environment, acting in the environment, and observing the environment) are running simultaneously without waiting for each other. Observing is also decoupled because whether an action would result in a positive reward can only be observed asynchronously after tens of minutes when the ad is clicked. Each of the three processes in our algorithm can be deployed to threads in multiple machines to improve runtime performance (Figure 3).

Though every auction in which we participate shares the same ads budgets and stock units, state transitions in the auction game are stochastic for the uncertainty of user activity. Under this consideration, our algorithm samples the next state of the $t$th auction from the set $(t, t + I_{transition}]$ where $I_{transition}$ is a hyper parameter of the algorithm.

Besides, different publishers always have very different CTR, CVR or ROI. Therefore, auctions from different publishers should be considered as different games. It's challenging for an agent to bid different auction games at the same time. However, training independent agents for different games as (Mnih et al. 2015) will make more states unobservable. Our solution is requiring the next state of the $t$th auction to be from the same publisher $P_t$.

**Data Augmentation and the Loss**

Assume we have a stochastic transition $(\phi_{t^-}, a_{t^-}, r_{t^-}, \phi_{t^+})$ as discussed above, considering the property of GSP auctions and the definition of $a_{t^-}$ and $r_{t^-}$, we have a deduction that any bid above $a_{t^-}$ would win the auction $t^-$ and any bid below $r_{t^-}$ would lose the auction. Given the deduction, for all $a \in A$, we redefine rewards of the auction $t^-$ as:

$$r_{t^-,a} := \begin{cases} 0 & \text{for all } a < a_{t^-} \\ r_{t^-} & \text{otherwise} \end{cases} \quad (2)$$

Combining Formula (1) and Formula (2) results in the following definition:

$$y_{t^-,a} := \begin{cases} r_{t^-,a} & \text{terminal } \phi_{t^+} \\ r_{t^-,a} + \gamma \max_{a'} Q(\phi_{t^+}, a'; \theta_T) & \text{otherwise} \end{cases} \quad (3)$$

And we define the loss function of LADDER as:

$$L_{LADDER}(\theta) = \frac{1}{C+1} \sum_{a=0}^{C} (y_{t,a} - Q(\phi_{t^-}, a; \theta))^2 \quad (4)$$

The original loss of DQN as Formula (1) still works, especially for application whose actions are not as correlated as auction games. Although we use DQN in this paper, Double DQN and Dueling Double DQN (Wang et al. 2015) can be naturally incorporated in out algorithm.

To maximize revenue while keeping breakeven, we introduce a weighted sampling method to tune the importance of positive rewards, which is controlled by the hyper parameter $\pi$. We also use an experience memory as in DQN. The full algorithm, which we call deep asynchronous stochastic Q-learning, is presented in Algorithm 1.

---

**Algorithm 1** Deep asynchronous stochastic Q-learning

Initialize experience memory $D$ to capacity $N$
Initialize parameters $(\theta, \theta_S, \theta_T)$ of action-value function $Q$ with random weights
**procedure** Serving
    **while** *true* **do**
        Get auction $A_t$ of publisher $P_t$ at timestamp $t$
        $\phi_t := \phi(A_t, Info_t,)$
        Asynchronously fetch snapshot of parameters $\theta$ to $\theta_S$
        With probability $\varepsilon$ select a random bid $a_t$
        otherwise select $a_t = max_a Q^*(\phi_t, a; \theta_S)$
        Respond to bidding request $A_t$ with bid $a_t$
    **end while**
**procedure** OBSERVING
    **while** *true* **do**
        Update $Info$
        Observe reward $r_{t'}$ and store $(\phi_{t'}, a_{t'}, r_{t'})$ in $D$
    **end while**
**procedure** TRAINING
    **while** *true* **do**
        Sample random mini-batch of stochastic transitions $(\phi_{t^-}, a_{t^-}, r_{t^-}, \phi_{t^+})$ from $D$ with probability:
$$prob_{t^-} = \begin{cases} \pi & \text{if } r_{t^-} < 0 \\ 1 & \text{otherwise} \end{cases}$$
        where $t^-$ and $t^+$ meet the constraints:
        $t^+ \sim U((t^-, t^- + I_{transition}])$ and $P_{t^+} = P_{t^-}$
        For all $a \in A$, perform:
$$y_{t^-,a} := \begin{cases} r_{t^-,a} & \text{terminal } \phi_{t^+} \\ r_{t^-,a} + \gamma \max_{a'} Q(\phi_{t^+}, a'; \theta_T) & \text{otherwise} \end{cases}$$
        Perform an SGD step on $\frac{1}{C+1}\sum_{a=0}^{C}(y_{t,a} - Q(\phi_{t^-}, a; \theta))^2$
        $\theta_T := \theta$ every $I_T$ steps
    **end while**
**main**
    Asynchronously start SERVING, OBSERVING and TRAINING

---

## Experimental Results

The experiments of LADDER were run on four important publishers that occupy a significant part of the revenues of JD.com's DSP business. We try to improve both revenue and profits of the publishers in the experiments.

### Experiment Setup

We run the training procedure of Algorithm 1 on 4 Tesla K80 GPUs and the serving procedure on 24 PC servers (Figure 3) with a high-performance C++ sparse convolver. We use RMSProp to optimize the loss in Formula (4) with a learning rate of $5^{-9}$. The usage of $\varepsilon$-greedy was restricted with an $\varepsilon$ of $1^{-3}$ to minimize negative influence on the business. Training was decomposed into 2 phases:

*Imitation*. We filled the experience memory with data generated by the ECPM policy of the hybrid system. At this stage, before enough self-generated data get into the memory, LADDER is just learning the ECPM policy. In this cold-starting phase, LADDER interacts little with the environment thus ensures that losses are under control.

*Introspection*. After several hours of imitation (the time actually required depends on $N$), LADDER starts to learn from data generated by its own policy.

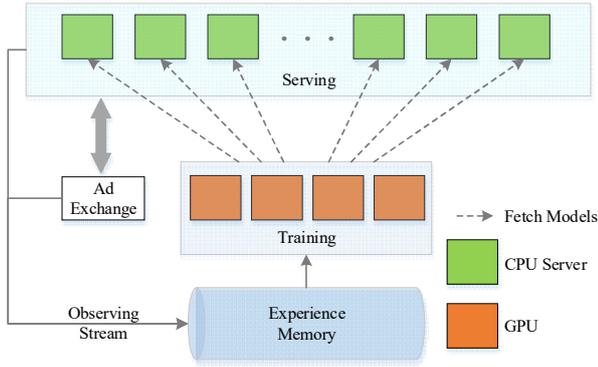

*Figure 3: An example of asynchronous deployment of LADDER*

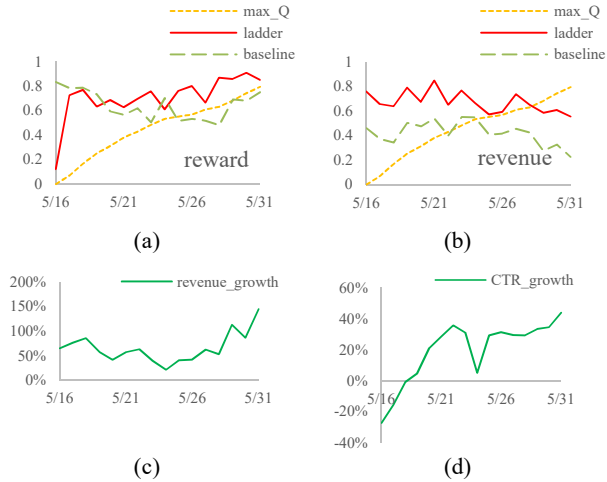

*Figures 4: Experiment results with the ECPM policy as baseline.*

### Evaluation

In May 2017, we evaluated LADDER with online A/B test in an overlapping experiment system similar to (Tang et al. 2010) and regarded the ECPM policy as baseline. In the beginning, LADDER was bidding 10% of the auctions per day and the remaining 90% was running the baseline. We launched LADDER at 90% of the auctions in the 8th day and keep the rest 10% as holdback which run the baseline policy for months for the sake of scientific rigor. The experiment system performed a proportional normalization to all experiments for ease of comparison.

Figures 4 shows the performance comparisons between LADDER and baseline. We normalized all data in the figures into range [0,1] for privacy. Figure 4(a) shows the rewards (profits) comparison, where we can see that LADDER incurred huge losses on the first day in the imitation phase because it tended to bid up all requests for exploration. It soon turned into the second phase and caught up with the baseline the next day, and eventually outperformed the baseline since day 5. Notice that the Q curve well fitted the curve of rewards of LADDER. There was a retreat at day 8 because we launched LADDER that day, therefore the experimental data were mixed up.

Figure 4(b) and Figure 4(c) shows the revenue growth: LADDER made a huge improvement by more than 50% since the first day. It seems that LADDER had learned the key of economies of scale that more revenues always generate more profits. Figure 4(d) shows that LADDER also raised CTR as much as about 35%, which is reasonable because the experimented publishers was on a CPC basis.

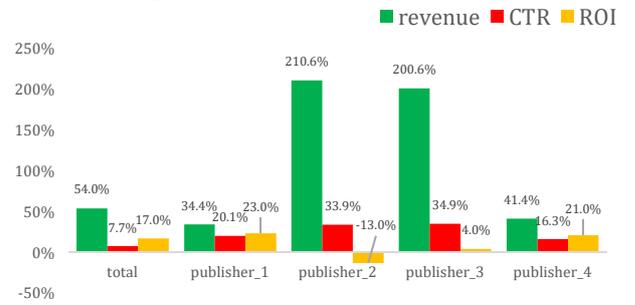

*Figure 5: Experimental data on JD's June 18$^{th}$ anniversary sale*

According to the holdback, the improvements are permanent. Especially, during JD.com's June 18th anniversary sale of 2017, LADDER increased the revenue of the 4 publishers by 54% and the advertisers ROI by 17% as shown in Figure 5, thus contributed a growth of 17% to the total revenue of JD's DSP business and 7% to the total ROI of the sale. The improvement during the sale proves the adaptability and responsiveness of LADDER in a highly volatile and competitive environment.

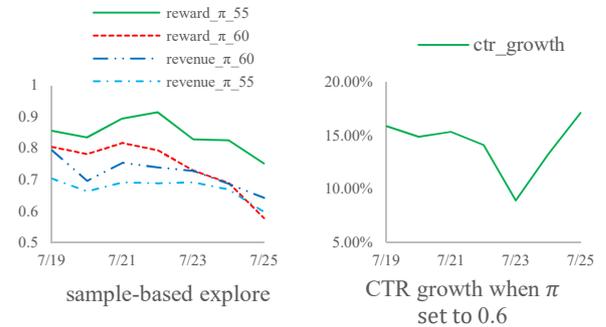

*Figures 6: The influence of weighted sampling.*

## Exploration and Exploitation

The hyper parameter $\pi$ controls the balance between exploration and exploitation. To maximize the revenues, our launched deployment set $\pi$ to 0.6. As Figures 6 depict, when we decrease $\pi$ from 0.6 to 0.55, revenue decrease while rewards and CTR increase, which means the agent tends to explore less aggressively.

## Visualization

In order to figure out LADDER's capability of understanding high-level semantic information embedded in the plain-text description inputs, we use t-SNE to visualize the outputs of the hidden layer. Our analysis is from two angles.

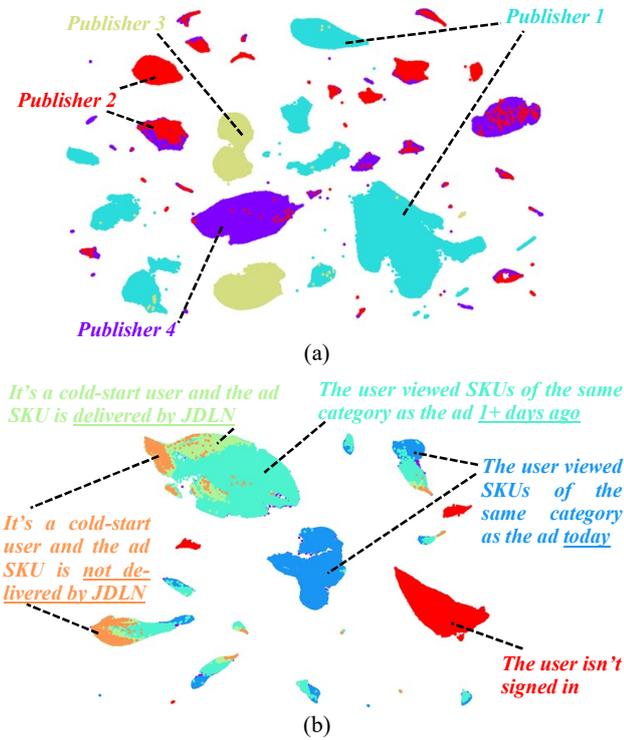

Figures 7: t-SNE visualization of LADDER. Different colors represent different publishers in (a) and different semantic in (b)

## Multiple Games in a Single Model

As mentioned earlier, LADDER serves very different publishers (aka different auction games) simultaneously. Although challenging, Figure 5 shows that LADDER had successfully learned from the plain-text inputs the difference of the publishers. It surpassed the baseline in revenue and CTR for each publisher evaluated. To technically verify how well LADDER can distinguish different publishers, we use publisher type as label and visualize 1000,000 random samples in Figure 7(a). As expected, all 4 publishers are mapped into separate clusters perfectly.

## Complex Semantic

In Figure 7(a), samples of the same publisher scatter into several clusters. In fact, LADDER's learned semantic much more complex than publisher IDs. For further analysis, we visualize data only of publisher 1 from the same 1000,000 samples. As Figure 7(b) shows, LADDER has learned rather complex conditions from the plain-text inputs, which are essential to bid an auction. E.g. SKUs delivered by JD logistic network (JDLN) may be more attractive for a cold-start user because JDLN is well-known to feature a superior user experience. As the left part of Figure 7(b) indicates, LADDER recognizes these situations.

## Future Work

Real-time online auctions are not the only large scale real world problems in which human-level agents excel. Considering that ADXs mimics stock exchanges, applying LADDER in quantitative trading is also of great interest and challenge.

Recommendation system is a domain with similarities to online advertising, so our approach should work in the area with a domain specific loss function.

What we are working on is applying LADDER not only for bidding but also in the ranking phase of online advertising, which may also bring significant business benefits.

## Conclusions

We present a reinforcement learning agent namely LADDER in this paper for solving the auction game of JD DSP. Our aim is to create a human-level agent that is capable of not only saving manpower while performing as well as or even better than humans, but also directly understanding the situation of an auction from a plain-text description, as human experts do. As the result, LADDER reach the goal by easily outperforming the existing industrial state-of-the-art solution in A/B tests, which means it has made full use the high-level semantic information in the auction game without sophisticated feature engineering and reacts to the changing auction environment immediately.

We also introduce DASQN, an asynchronous stochastic Q-network which totally decouples the learning, observing and acting processes in Q-learning, hence greatly improving its run-time performance and enabling the algorithm to solve large scale real-world problems.